\documentclass[12pt,reqno]{amsart}
\usepackage{latexsym, amsmath, amscd, amssymb, amsthm,longtable} 
\usepackage[T2A]{fontenc}
 \usepackage[cp1251]{inputenc}
\usepackage[english]{babel}
\newtheorem{definition}{Definition}

\newtheorem*{proposition}{Theorem}

\usepackage{graphicx}

\usepackage{url}

\usepackage{float}
\begin{document}

\title[Moment-based Scale-Invariant Quality Measure]
 {
MSIQ: Moment-based Scale-Invariant Quality Measure
  for Single Image Super-Resolution
}
\author{Leonid Bedratyuk}
\address{ Khmelnytsky National University, Faculty of Information Technology, Ukraine}
\email{leonidbedratyuk@khmnu.edu.ua}
\maketitle

\begin{abstract}

Assessing the quality of single image super-resolution (SISR) results remains an open methodological problem. Common full-reference metrics (PSNR, SSIM, LPIPS) do not explicitly evaluate the preservation of the geometric structure of images, which is critical for the correctness of scale-based reconstruction. In addition, they require the forced alignment of images to the same size (\textit{forced resizing}), which introduces an external interpolation error into the evaluation process.

This paper proposes a diagnostic scale-invariant quality measure, MSIQ (\textit{Moment-based Scale-Invariant Quality}), based on the comparison of normalized central geometric moments of two images. MSIQ enables direct comparison of images with different spatial resolutions without resizing, is mathematically deterministic (\textit{model-free}), and has an analytical form. To provide a theoretical basis for the approach, we introduce a conceptual distinction between the ability of metrics to monotonically track degradation (\textit{tracking ability}) and their geometric selectivity (\textit{geometric specificity}).

The experimental validation confirmed the stability of MSIQ under uniform scaling and, at the same time, revealed the high sensitivity of traditional metrics to the choice of interpolation method. The results show that MSIQ has pronounced geometric selectivity: the proposed measure effectively separates geometric deformations from non-geometric artifacts, in particular JPEG compression, unlike pixel-based and perceptual metrics. It is also shown that the response of MSIQ to structural perturbations remains stable across different classes of SR algorithms, including DNN models with different architectures. The proposed measure is a complementary diagnostic tool for domains where geometric fidelity has priority, in particular medical imaging and remote sensing.

\vspace{0.5cm}

\end{abstract}

\section{Introduction}

Single Image Super-Resolution is the task of reconstructing a high-resolution
image \(I_{SR}\) from a single low-resolution input image \(I_{LR}\), and it is
one of the central directions in modern computer vision~\cite{Yue2016, Chen2022}.
Deep learning has substantially improved the quality of SR results: modern
CNN-, Transformer-, and diffusion-based approaches achieve significantly higher
PSNR and SSIM values than classical interpolation methods~\cite{Lugmayr2022, AKS2022, Gao2023}.
At the same time, the increasing capabilities of SR models have sharpened the
problem of their objective evaluation: it is well known that rankings based on
standard indicators do not always agree with subjective quality assessments~\cite{Reibman2006, Ma2017}.

The image quality assessment literature has repeatedly emphasized that classical
objective metrics have substantial limitations. MSE and PSNR primarily evaluate
pixel-wise error and do not take into account either image content or the
properties of human visual perception~\cite{Engeldrum2001, Ridder2002}.
Metrics based on models of the visual system are closer to subjective assessment,
but they depend on complex and incomplete models of perception~\cite{Bianco2009}.
Even SSIM, which accounts for structural similarity~\cite{Wang2004}, may be
unstable with respect to different types of distortions. Moreover, most
full-reference metrics, including MS-SSIM, LPIPS, and DISTS, require the
reference and test images to have the same size~\cite{Wang2004, Sheikh2006, Ding2022DISTS}.
This makes them unsuitable for direct comparison of data with different
resolutions without an additional size-matching step, since forced resizing
introduces an external interpolation error into the evaluation process. As noted
in general approaches to data quality assessment, the use of standard
full-reference comparisons after geometric transformations requires particular
care, because resizing is not a property of the SR model itself, but it can
critically affect the result.

Thus, a substantial part of the evaluation may depend on the methodological
choice of the comparison procedure, rather than only on the quality of the
reconstruction itself. This is especially critical because existing metrics do
not explicitly evaluate the correctness of the scale transformation. SISR is,
by its nature, a transition between two scales of spatial discretization; hence,
a high-quality result should preserve properties that are invariant under
uniform scaling. If anisotropic deformations, changes in object proportions, or
perspective distortions arise during reconstruction, none of the classical
metrics is designed to detect this particular class of errors.

The problem addressed in this paper can be formulated as follows:
\emph{can one construct a scale-invariant measure that explicitly evaluates
the preservation of the geometric structure of an SR image with respect to the
reference image \(I_{GT}\), detects geometric anomalies in the super-resolution
result, and does not require forced resizing?}

To solve this problem, we propose to use normalized central geometric moments,
a classical tool for shape analysis and pattern recognition~\cite{hu1962, Flusser2017}.
Their key property is invariance under uniform scaling and translation in the
continuous model. Thus, in the discrete case, a geometrically correct SR result
should preserve the moment descriptor of the reference image up to discretization
and interpolation errors; therefore, the corresponding normalized central
geometric moments of \(I_{GT}\) and \(I_{SR}\) should be close. The deviation
between such descriptors is naturally interpreted as a measure of violation of
scale-invariant consistency. This makes it possible to avoid forced resizing
by comparing images directly in the space of moment descriptors.

Based on these observations, we propose MSIQ (\emph{Moment-based Scale-Invariant Quality}),
a measure of discrepancy between descriptors formed by normalized central
geometric moments of the reference and reconstructed images. MSIQ is considered
as a full-reference measure for SISR: the reference is the HR ground-truth image
\(I_{GT}\), and the test image is the super-resolution result \(I_{SR}\). The
difference between MSIQ and standard full-reference metrics is that the
comparison is performed not in pixel space, but in the space of normalized
moment descriptors.

MSIQ does not replace general-purpose indicators such as PSNR, SSIM, and LPIPS;
rather, it plays a complementary diagnostic role. It enables the detection of
violations of global geometry that are usually averaged out in pixel-wise error
measures or in the analysis of local perceptual features. Within the FUN
paradigm, where traditional metrics focus on signal fidelity and generative
approaches focus on naturalness, MSIQ fills the niche of objective control of
structural integrity. This is critically important for scientific and technical
visualization, where geometric correctness is a priority even when the image has
high subjective quality.

The proposed study integrates methods that are foundational for Pattern
Recognition: the mathematical apparatus of moment invariants for the analysis
of geometric structures~\cite{Flusser2017, hu1962}, computational assessment of
visual data quality~\cite{Ma2017, Zhou2022, Lyapustin2022}, and comparative
analysis of image descriptors in feature spaces~\cite{wang2012, premaratne2012}.
At the same time, the paper addresses the current problem of verifying SISR
results, which has become especially important due to the spread of generative
models~\cite{Lugmayr2022, Saharia2022, Wu2024, Wang2024} that may introduce
unpredictable structural deformations.

The proposed approach, namely evaluating the similarity of objects through
their invariant representations, fully conforms to the Pattern Recognition
paradigm, in which the identification of patterns and the metrization of their
similarity are based on extracting invariant features that describe the geometry
of the image scene independently of its discretization conditions or scale.

The contributions of this work can be summarized in four main points:
\begin{enumerate}
\item A conceptual approach to SISR evaluation is proposed through verification
of the integrity of scale-invariant moment descriptors of images. This approach
transfers the comparison of GT and SR data into a feature space, thereby
eliminating the methodological dependence on forced resizing and the associated
interpolation error.

\item This approach is implemented as the analytical measure MSIQ, which uses
the distance between descriptors as an indicator of the stability of the
moment-geometric structure of the image under scale transformations. MSIQ is
a model-free measure and does not require a forward pass through a pretrained
neural network, unlike LPIPS and DISTS.

\item It is experimentally shown that \textit{MSIQ} has high geometric
selectivity, or \textit{geometric specificity}: it can isolate geometric
deformations from non-geometric artifacts, in particular JPEG compression. This
makes it possible to detect critical structural anomalies and violations of
proportions even under small perturbations in the outputs of SR models, where
traditional pixel-based and perceptual metrics show low selectivity.

\item A methodological distinction is introduced between two characteristics of
quality indicators: the ability to monotonically track degradation, or
\textit{tracking ability}, and geometric specificity. This distinction
emphasizes that a monotonic response of a quality indicator to an increase in
distortion strength does not yet imply its selectivity with respect to the
geometric nature of that distortion.
\end{enumerate}

In this context, MSIQ is positioned as a specialized diagnostic tool specifically
for assessing geometric fidelity.

The structure of the paper is organized as follows. Section~2 presents a
critical review of the literature, covering the evolution of metrics from
pixel-based to perceptual ones, the analysis of specialized SR-IQA approaches,
and the use of moment invariants, on the basis of which the existing research
gap is clearly identified. Section~3 gives the formal definition of the
\textit{MSIQ} measure and discusses its main properties. Section~4 describes
the experimental protocol, and Section~5 presents the validation results,
covering four series of controlled tests on stability and selectivity. Section~6
provides a critical discussion of the obtained results, and Section~7 summarizes
the conclusions and outlines promising directions for future research.

\section{Related Work}
\label{sec:related}

This section systematizes approaches to assessing the quality of SISR results
and defines the position of MSIQ among existing methods.
The discussion covers four aspects: pixel-based, perceptual, and specialized
SR-IQA metrics; the problem of evaluation under geometric mismatch between
images; moment invariants in image analysis; and the resulting research gap.

\subsection{Pixel-based, perceptual, and specialized SR-IQA metrics}

The evaluation of SISR results traditionally relies on full-reference
metrics such as PSNR and SSIM~\cite{Wang2004, Yue2016, Arabboev2024}.
PSNR measures the mean squared error in pixel space;
SSIM accounts for local similarity in luminance, contrast, and structure.
Both metrics are computationally simple and have a well-established
interpretation. At the same time, they are not designed for the explicit
diagnosis of the geometric correctness of scale-based reconstruction.
Moreover, they require the compared images to have the same spatial size,
which prevents direct comparison of \(I_{GT}\) and \(I_{SR}\) without
forced resizing, a procedure that itself introduces interpolation error
and is not a property of the evaluated model~\cite{Batini2016}.

To bridge the gap between objective indicators and subjective perception,
perceptual metrics such as LPIPS, DISTS, FID, and NIQE are widely used
~\cite{Zhang2018, Ding2022DISTS, Chen2022, Wu2024, Wang2024}.
They agree better with visual assessments, especially for generative and
diffusion-based SR models. However, as Blau and Michaeli~\cite{Blau2018}
theoretically showed, maximizing perceptual quality inevitably leads to
an increase in structural deviations from the reference. Since LPIPS and
similar metrics rely on features of pretrained networks, they are prone
to texture bias~\cite{Zhang2018} and may fail to detect violations of the
global geometry of objects. In domains where geometric fidelity is a
critical requirement, in particular medical imaging and remote sensing,
perceptual metrics are insufficient without specialized criteria
~\cite{Jiang2022, Lyapustin2022, Zhou2022}.

Recognition of the limitations of general-purpose indicators has stimulated
the development of specialized SR-IQA methods. Reibman et al.~\cite{Reibman2006}
were among the first to systematically describe the gap between objective
criteria and subjective perception in the presence of aliasing and ringing.
Ma et al.~\cite{Ma2017} proposed a no-reference metric for SISR based on
local frequency and global statistical features. Jiang et al.~\cite{Jiang2022}
introduced KLTSRQA and the RealSRQ dataset. Greeshma and Bindu~\cite{Greeshma2020}
proposed the Super-resolution Quality Criterion (SRQC), which combines gradient
and multiscale features for assessing the quality of SR images. Zhou and
Wang~\cite{Zhou2022} explicitly distinguish deterministic fidelity and
statistical fidelity in the SRIF metric. Lyapustin et al.~\cite{Lyapustin2022}
proposed ERQA, a metric for edge-structure restoration that accounts for local
gradient shifts. Fan et al.~\cite{Fan2025} consider no-reference SR-IQA based
on contrastive learning with awareness of the upscaling factor, which emphasizes
the importance of scale itself in SISR evaluation tasks. The survey by Shu
et al.~\cite{Shu2025} systematizes the current state of the field. Among
recent trends, it is worth noting RQI, which accounts for imperfections of
the GT images themselves~\cite{Su2025RQI}; the self-supervised S\(^{3}\)RIQA
methodology~\cite{Majlessi2026}; and the use of NR-IQA predictors directly
in loss functions when training SR models~\cite{Zhang2025IQA}. These works
mainly focus on perceptual quality, generative artifacts, or statistical
naturalness; the issue of scale-invariant geometric fidelity remains secondary
in them or is not formulated explicitly.

\subsection{Evaluation under geometric mismatch between images}

Standard full-reference metrics belong to aligned-reference approaches:
they assume that \(I_{GT}\) and \(I_{SR}\) have the same pixel grid and
can be compared directly. When images of different sizes are compared,
forced alignment is required; however, this procedure is not neutral:
the choice of interpolation method substantially affects the numerical
values of PSNR and SSIM without being related to the quality of the model
itself~\cite{Batini2016}.

The GDR-IQA direction, \textit{Geometrically-Disparate-Reference IQA},
addresses the evaluation problem when the reference and test images are
geometrically mismatched. DeepSSIM~\cite{Zhang2026DeepSSIM} replaces
direct pixel-wise comparison with evaluation through deep features, which
provides robustness to geometric discrepancies. MSIQ works in a closely
related methodological setting, but uses a different apparatus: instead
of learned deep features, it relies on an analytical model-free moment
descriptor that is invariant to translation and uniform scaling.

The class of structure-aware metrics that are partially distinct from MSIQ
includes HaarPSI~\cite{Reisenhofer2018HaarPSI}, MDSI~\cite{Nafchi2016MDSI},
and GMSD~\cite{Xue2014GMSD}. These methods evaluate similarity through
local Haar wavelet coefficients, gradient maps, or deviation pooling and
are designed for comparing images on the same pixel grid. MSIQ differs
from them in that it operates with global scale-invariant moment descriptors
and therefore does not require preliminary size alignment.

\subsection{Moment invariants in image analysis}

Moment invariants are a classical tool in shape analysis, pattern recognition,
and computer vision. Hu~\cite{hu1962} found certain polynomial combinations
of normalized central moments that are invariant to translation, rotation,
and uniform scaling. Further development of the theory is associated with
the works of Flusser and Suk, in particular on affine moment invariants
~\cite{flusser1993} and moment-based approaches to image registration
~\cite{flusser1994}. A modern systematic exposition of moment analysis is
given in the monograph by Flusser, Suk, and Zitov\'a~\cite{Flusser2017}.

In the context of image quality assessment, moment descriptors have also
been used before. In particular, they have been used for blur assessment
~\cite{wee2007}, for constructing IQA metrics based on discrete orthogonal
moments~\cite{wee2010}, for image comparison through invariant moment
descriptions~\cite{wang2012}, and for moment descriptors in the wavelet
domain~\cite{premaratne2012,premaratne2014}. These works are important
because they demonstrate the general principle that images can be compared
not directly in pixel space, but through compact or structured descriptors.

In super-resolution tasks, moments have previously been used mainly as a
tool for reconstruction or registration, rather than as an independent
quality measure. Singh and Aggarwal~\cite{singh2017} used rotation-invariant
orthogonal invariants for SISR reconstruction, while Kanan et al.~\cite{rashidy2015}
used pseudo-Zernike moments for multi-frame super-resolution. These approaches
confirm the relevance of the moment apparatus for SR, but they do not formulate
the problem of assessing the geometric correctness of the SR result.

The works closest to our formulation are those of Wang~\cite{wang2012}
and Premaratne~\cite{premaratne2014}, where image quality or similarity is
assessed through moment invariants, in particular at the level of wavelet
approximations. These approaches are important because they demonstrate the
possibility of comparing images through invariant descriptors rather than
directly through pixel-wise error.

A direct precursor of the present work is our previous paper~\cite{Bedratyuk2020},
in which normalized central moments were used to assess the quality of classical
image scaling.

The invariants used in MSIQ differ fundamentally from the classical Hu
invariants. Hu invariants are simultaneously invariant to translation, uniform
scaling, and rotation. For the SISR task, rotation, shear, perspective distortion,
and anisotropic scaling are geometric pathologies that the metric must detect,
rather than admissible transformations that it should ignore. In addition,
the seven Hu invariants do not form an algebraically independent complete
system~\cite{Flusser2017, Bed2020} and are insufficient for diagnosing subtle
geometric changes. Instead, MSIQ relies on normalized central geometric moments,
which provide invariance only to translation and uniform scaling, while preserving
sensitivity to other geometric deformations. It is precisely this
\emph{selective invariance} that corresponds to the nature of the SISR task.

\subsection{Research gap}

The systematization of existing approaches reveals the following picture.
Classical full-reference metrics (PSNR, SSIM) depend on forced resizing and
pixel alignment. Perceptual metrics (LPIPS, DISTS) measure visual naturalness,
but do not have a direct geometric interpretation. Specialized SR-IQA approaches
(SRIF, KLTSRQA, ERQA, S\(^{3}\)RIQA) account for the specificity of SR artifacts,
but they focus on local features or perceptual quality without verifying the
preservation of a global scale-invariant moment description. Finally, GDR-IQA
approaches (DeepSSIM) address the problem of geometric mismatch by means of
learned features rather than an analytical model-free descriptor.

The question that remains undeveloped is natural precisely for SISR:
\emph{does the SR result preserve the global characteristics of the reference
image \(I_{GT}\) that are invariant under uniform scaling?} If anisotropic
deformations, shear, or violations of proportions arise during reconstruction,
the scale-invariant consistency between \(I_{GT}\) and \(I_{SR}\) is violated,
but none of the existing metrics tests this property explicitly.

MSIQ addresses this gap by comparing \(I_{GT}\) and \(I_{SR}\) in the space of
normalized central geometric moments and measuring the distance between their
scale-invariant descriptors. Thus, MSIQ is a \emph{model-free},
\emph{scale-invariant}, and \emph{geometry-specific} complement to existing
metrics, a diagnostic criterion for assessing the global geometric fidelity of
SISR reconstruction that complements PSNR, SSIM, LPIPS, DISTS, and ERQA.

Although normalized central moments provide invariance to both translation and
uniform scaling, the name MSIQ emphasizes specifically the \emph{scale-invariant}
property, because scaling is the central transformation in the SISR task.
\section{The Proposed MSIQ Measure}
\label{sec:theory}

This section presents the mathematical construction of MSIQ.
First, geometric moments and their scale-invariant normalizations are introduced;
then the moment descriptor of an image is formed, and finally MSIQ is defined
as a distance between the descriptors of the $I_{GT}$ and $I_{SR}$ images.

\subsection{Geometric moments and their scale-invariant normalizations}

For a continuous grayscale image
$f\colon\Omega\subset\mathbb{R}^{2}\to\mathbb{R}$,
the geometric moment of order $(p,q)$ is defined by~\cite{hu1962}:
\begin{gather}
\label{eq:moment}
    m_{pq}(f)
    =
    \iint_{\Omega} x^{p}\,y^{q}\,f(x,y)\,dx\,dy,
    \qquad p,q \geq 0.
\end{gather}
For a digital image defined on the discrete grid
$\{0,\ldots,H-1\}\times\{0,\ldots,W-1\}$, the integral is replaced by
the corresponding sum:
$$
    m_{pq}(I)
    =
    \sum_{i=0}^{H-1}\sum_{j=0}^{W-1}
    i^{p}\,j^{q}\,I(i,j).
$$
The moment $m_{00}$ represents the total image intensity, or its mass,
while the centroid coordinates
$$
    \bar{x} = \frac{m_{10}}{m_{00}},
    \qquad
    \bar{y} = \frac{m_{01}}{m_{00}}
$$
define the center of mass of the image.

Geometric moments are not invariant to translation.
Central geometric moments
$$
    \mu_{pq}(f)
    =
    \iint_{\Omega}
    (x-\bar{x})^{p}\,(y-\bar{y})^{q}\,f(x,y)\,dx\,dy
$$
are invariant to translation; in particular, $\mu_{10}=\mu_{01}=0$ and
$\mu_{00}=m_{00}$.

To provide additional invariance with respect to uniform scaling, normalized
central geometric moments are introduced:
\begin{gather}
\label{eq:nu}
    \nu_{pq}(f)
    =
    \frac{\mu_{pq}(f)}{\mu_{00}(f)^{\,1+\frac{p+q}{2}}},
    \qquad
    \mu_{00}(f)>0.
\end{gather}

\begin{proposition}[\cite{hu1962},\cite{Flusser2017}]\label{prop:invariance}
In the continuous model, the normalized central moments $\nu_{pq}$ are
invariant to translation and uniform scaling of the image.
\end{proposition}

In the discrete case, small residual deviations are expected due to resampling
and numerical computation of moments; their magnitude is experimentally assessed
in §\ref{sec:exp1}.

\subsection{Moment descriptor}

In the SISR task, the images $I_{GT}$ and $I_{SR}$ represent the same scene
geometry at different levels of spatial discretization.
Therefore, a geometrically correct SR result should preserve scale-invariant
moment features:
\begin{gather}
\label{eq:consistency}
    \nu_{pq}(I_{GT}) \approx \nu_{pq}(I_{SR}).
\end{gather}
Deviation from this condition is naturally interpreted as a measure of violation
of scale-invariant geometric consistency between $I_{GT}$ and $I_{SR}$.

For a fixed order $N$, we use the triangular domain of moment indices
$$
    \Omega_N^{\triangle}
    =
    \bigl\{(p,q)\colon p,q\geq 0,\;p+q\leq N\bigr\},
$$
which standardly restricts the total order of moments.
Since the components $\nu_{00}=1$, $\nu_{10}=0$, and $\nu_{01}=0$ are
trivial and do not carry structural information, the informative subset is used
in the computations:
$$
    \Omega_N^{\mathrm{inf}}
    =
    \Omega_N^{\triangle}
    \setminus
    \bigl\{(0,0),(1,0),(0,1)\bigr\}.
$$

The \textbf{moment descriptor} of an image $I$ is defined as the ordered set of
normalized central moments:
\begin{gather}
\label{eq:descriptor}
    \mathcal{M}_{N}(I)
    =
    \bigl(\nu_{pq}(I)\bigr)_{(p,q)\in\Omega_N^{\mathrm{inf}}}.
\end{gather}

\subsection{Definition and variants of MSIQ}

\begin{definition}\label{def:msiq}
\textbf{MSIQ} (\emph{Moment-based Scale-Invariant Quality}) is a measure of
discrepancy between the moment descriptors of the reference and SR images:
\begin{gather}
\label{eq:msiq_general}
    \operatorname{MSIQ}_{N}(I_{GT},I_{SR})
    =
    d\bigl(\mathcal{M}_{N}(I_{GT}),\,\mathcal{M}_{N}(I_{SR})\bigr),
\end{gather}
where $d$ is the chosen distance function.
\end{definition}

This paper uses two computational versions of MSIQ.

\noindent\textbf{Baseline RMSE version:}
\begin{gather*}
    \operatorname{MSIQ}_{RMSE}(I_{GT},I_{SR})
    =
    \left(
        \frac{1}{|\Omega_N^{\mathrm{inf}}|}
        \sum_{(p,q)\in\Omega_N^{\mathrm{inf}}}
        \bigl(\nu_{pq}(I_{GT})-\nu_{pq}(I_{SR})\bigr)^{2}
    \right)^{1/2}.
\end{gather*}
\noindent\textbf{Weighted version:}
\begin{gather*}
    \operatorname{MSIQ}_{W}(I_{GT},I_{SR})
    =
    \left(
        \sum_{(p,q)\in\Omega_N^{\mathrm{inf}}}
        w_{pq}\,
        \bigl(\nu_{pq}(I_{GT})-\nu_{pq}(I_{SR})\bigr)^{2}
    \right)^{1/2},
\end{gather*}
where the weights
$$
    w_{pq}=\frac{1}{1+p+q}
$$
are inversely proportional to the order of the moment. This weighting reduces
the contribution of higher-order moments, which may be more sensitive to
discretization and numerical errors.

Unlike $\operatorname{MSIQ}_{RMSE}$, the weighted version
$\operatorname{MSIQ}_{W}$ is not normalized by the sum of weights. Therefore,
its absolute values should not be directly compared with the values of
$\operatorname{MSIQ}_{RMSE}$. In this paper, $\operatorname{MSIQ}_{W}$ is used
as an auxiliary robust variant for checking the stability of conclusions
regarding the geometric specificity of MSIQ.

A small value of $\operatorname{MSIQ}_{RMSE}$ or $\operatorname{MSIQ}_{W}$
indicates preservation of the scale-invariant geometric structure of the image.
Large values indicate the presence of global geometric violations: anisotropic
scaling, shear deformation, rotation, perspective distortion, or violation of
proportions.

In this paper, the baseline protocol is fixed as
$$
    N=4,
    \quad
    \Omega_N=\Omega_4^{\triangle},
    \quad
    |\Omega_4^{\triangle}|=15,
    \quad
    |\Omega_4^{\mathrm{inf}}|=12.
$$
Both versions of MSIQ are computed using 12 nontrivial normalized central
moments. Hereafter, the notation $\operatorname{MSIQ}_{RMSE}$ and
$\operatorname{MSIQ}_{W}$ refers to this fixed protocol unless stated otherwise.

\subsection{Protocol for color images}

In the main experiments, color images were first converted to a single-channel
grayscale representation, after which MSIQ was computed for the corresponding
scalar intensity function. This makes it possible to apply the same moment
apparatus as for monochrome images and to focus the evaluation on preserving
the scale-invariant geometric structure without mixing it with color
differences. Extending MSIQ to channel-wise processing of color images is beyond
the scope of this paper.

\subsection{Properties and limitations}

The following three properties define the role of MSIQ as a specialized
diagnostic tool for assessing geometric correctness in SISR tasks:

\begin{enumerate}
    \item \emph{Resizing-free}: MSIQ is computed directly for
    $I_{GT}$ and $I_{SR}$ of different sizes without forced resizing,
    since the comparison is performed in the space of moment descriptors.

    \item \emph{Model-free}: the measure is defined analytically and does not
    depend on a pretrained neural network, unlike LPIPS and DISTS.

    \item \emph{Geometry-specific}: MSIQ responds much more strongly to
    geometric deformations (anisotropic scaling, shear, rotation, and
    perspective distortion) than to non-geometric artifacts, in particular
    JPEG compression.
\end{enumerate}

The first two properties are direct consequences of the mathematical
construction of the measure; the third is confirmed experimentally
(§\ref{sec:results_exp2}).

\textbf{Limitations of the measure:}

\begin{itemize}
    \item \textit{Global nature of the assessment.}
    As an integral measure, MSIQ responds weakly to small local texture artifacts
    if they do not shift the overall mass distribution. It is therefore advisable
    to use it in synergy with local or edge-based metrics, such as
    ERQA~\cite{Lyapustin2022}.

    \item \textit{Sensitivity to background.}
    Global moments describe the composition of the entire scene as a whole.
    For tasks in which the object of interest occupies only a small part of the
    frame, the use of a mask or ROI (region of interest) is recommended.

    \item \textit{Photometric non-invariance.}
    MSIQ is not invariant to global photometric transformations, such as changes
    in brightness, contrast, or gamma. In tasks where photometric variations are
    not part of the degradation being evaluated, preliminary intensity
    normalization or histogram matching may be required before computing MSIQ.

    \item \textit{Specialization.}
    MSIQ is not a universal perceptual metric. Its purpose is strictly
    diagnostic: verification of the scale-invariant geometric integrity of the
    SR result.
\end{itemize}

\section{Experiments}
\label{sec:experiments}

The experimental validation of MSIQ includes four interrelated
series of tests. The first tests the preservation of the scale-free
property in the discrete domain and demonstrates the instability of
standard FR metrics under forced resizing. The second analyzes the
ability of MSIQ to separate geometric deformations from non-geometric
artifacts on controlled model distortions. The third applies the same
protocol to the outputs of real SISR algorithms from different
architectural classes. The fourth confirms the stability of the
observed properties on standard SISR benchmarks, namely Set5, Set14,
and Urban100.

In all main experiments, a full-reference protocol is used:
$$
\operatorname{MSIQ}_{RMSE}(I_{GT},I_{SR}), 
\qquad 
\operatorname{MSIQ}_{W}(I_{GT},I_{SR}),
$$
where $I_{GT}$ is the HR ground-truth image and $I_{SR}$ is the
output of the SISR model.

\subsection{First experiment: validation of the scale-invariant property}

The aim of the first experiment is to quantitatively assess the
stability of MSIQ under changes in the spatial resolution of an image
and to demonstrate the methodological vulnerability of PSNR and SSIM
to the choice of the forced resizing method.

For each test image, a set of scaled copies is generated. The comparison
is implemented according to two schemes.
\begin{enumerate}
    \item \textit{For MSIQ:} the distance $\operatorname{MSIQ}_{RMSE}$
    is computed between the moment descriptor of the original image and
    that of its scaled copy directly in their original resolutions,
    without forced resizing.
    \item \textit{For PSNR and SSIM:} the scaled copy is forcibly
    returned to the original size using a set of standard interpolation
    methods; the range of the obtained values is interpreted as a
    measure of methodological instability.
\end{enumerate}

The experiment was conducted on six images: \texttt{camera},
\texttt{moon}, \texttt{coins}, \texttt{page}, \texttt{astronaut},
and \texttt{chelsea}.

\begin{itemize}
    \item \textit{Scaling:} scale factors
    $s\in\{0.5,\,0.75,\,1.5,\,2.0,\,3.0\}$ and five interpolators
    (\texttt{area}, \texttt{bilinear}, \texttt{lanczos4},
    \texttt{bicubic}, \texttt{nearest}) are used, giving 25 comparison
    pairs per image.
    \item \textit{MSIQ parameters:} $N=4$,
    $\Omega_N=\Omega_4^{\triangle}$, $|\Omega_4^{\mathrm{inf}}|=12$
    according to the protocol in §\ref{sec:theory}.
    \item \textit{FR metric parameters:} for forced resizing in the
    computation of PSNR and SSIM, the methods \texttt{nearest},
    \texttt{bilinear}, \texttt{bicubic}, and \texttt{lanczos4} are used.
    The \texttt{area} method is used only to generate the scaled copies,
    but not as a method for returning them to the original resolution.
\end{itemize}

To verify the correctness of the implementation, the following conditions
were checked: (1) the trivial moments $\nu_{00}\approx 1$,
$\nu_{10}\approx 0$, $\nu_{01}\approx 0$ at all iterations; and
(2) the condition $\operatorname{MSIQ}_{RMSE}(I,I)=0$ for each test image.

\subsection{Second experiment: geometric specificity of MSIQ}
\label{sec:exp2_protocol}

\subsubsection{Types of model distortions}

The aim of the second experiment is to test the ability of MSIQ to
separate geometric deformations from non-geometric artifacts according
to two criteria: tracking ability and geometric specificity. Five types
of degradation are used with the parameter
$\lambda\in\{0,\,0.05,\,0.10,\,0.15,\,0.20\}$:

\begin{enumerate}
    \item \textit{Anisotropic scaling (area-preserving):}
    $s_x=1+\lambda$, $s_y=1/(1+\lambda)$; this changes object
    proportions while preserving area.
    \item \textit{Shear deformation:} a shear transformation that
    violates the orthogonality of structures.
    \item \textit{Rotation:} rotation that changes orientation and
    the structure of odd-order moments.
    \item \textit{Perspective distortion:} a projective transformation
    that models a change in viewing angle.
    \item \textit{JPEG compression (negative control):}
    a non-geometric artifact that does not change the geometry of
    objects; it is used to test geometric specificity.
\end{enumerate}

\subsubsection{Metrics and the coefficient of geometric specificity}

For comparison, several groups of metrics are selected: classical
metrics (PSNR, SSIM), perceptual metrics (LPIPS, DISTS), and the
structure-oriented ERQA. To quantify geometric specificity, the following
coefficient is introduced:
\begin{equation}\label{eq:R}
    R_{M}(\lambda)
    =
    \frac{
        \operatorname{mean}_{d\in\mathcal{D}_{\mathrm{geom}}}
        \Delta M_{d}(\lambda)
    }{
        \Delta M_{\mathrm{JPEG}}(\lambda)
    },
\end{equation}
where
$$
\mathcal{D}_{\mathrm{geom}}
=
\{\text{anisotropic},\,\text{shear},\,\text{rotation},\,\text{perspective}\},
$$
and the metric increment is defined as
$$
    \Delta M_{d}(\lambda)
    =
    \begin{cases}
        M_{d}(\lambda)-M_{d}(0), & \text{lower-is-better,}\\
        M_{d}(0)-M_{d}(\lambda), & \text{higher-is-better.}
    \end{cases}
$$
A high value of $R_M(\lambda)$ indicates that the metric responds much
more strongly to geometric deformations than to the JPEG control.
The coefficient is computed for $\lambda=0.20$ (maximum degradation)
and $\lambda=0.05$ (weak degradation).

\subsubsection{Technical implementation}

The transformations are implemented using OpenCV with bicubic
interpolation to minimize external artifacts; affine transformations
are centered with respect to the geometric center of the image.
The parameters of perspective distortion are defined by shifts of
corner points proportional to $\lambda$. The JPEG quality parameter is
$$
    q = \max(1,\; 100 - 80\lambda),
$$
which provides a gradual increase in degradation from $q=100$ to $q=84$
at $\lambda=0.20$; this range corresponds to a moderate level of
JPEG compression and is chosen as a conservative control influence for
testing geometric specificity.

The experiment is conducted on the same set of six test images as
Experiment~1.

\subsection{Third experiment: controlled degradation on SISR outputs}

\subsubsection{Design and processing pipeline}

The third experiment tests the behavior of the metrics on the outputs
of real SISR algorithms. To avoid ambiguity regarding the ``reference''
geometric correctness of DNN outputs, a controlled perturbation protocol
is used: a degradation of known strength $\lambda$ is added to the
stabilized output of an SISR algorithm.

The processing pipeline for each image $I_{GT}$ is
\begin{equation}\label{eq:pipeline}
    I_{GT}
    \xrightarrow{\text{downsampling}}
    I_{LR}
    \xrightarrow{\text{SISR}}
    I_{SR}
    \xrightarrow{T_{\lambda}}
    T_{\lambda}(I_{SR}).
\end{equation}
The object of evaluation is the pair
$\bigl(I_{GT},\; T_{\lambda}(I_{SR})\bigr)$.
This protocol makes it possible to isolate the influence of the inserted
geometric pathology from the background artifacts of reconstruction.

\subsubsection{Methods under study and types of distortions}

Two classes of algorithms are used:
\begin{itemize}
    \item \textit{Classical interpolators:}
    \texttt{nearest}, \texttt{bilinear}, \texttt{bicubic},
    \texttt{lanczos4}.
    \item \textit{DNN models:}
    \texttt{FSRCNN}~\cite{Dong2016},
    \texttt{ESPCN}~\cite{Shi2016},
    \texttt{LapSRN}~\cite{Lai2017},
    \texttt{EDSR}~\cite{Lim2017}.
\end{itemize}
The scaling factors are $\times 2$, $\times 3$, and $\times 4$.
The types of degradation and the values of $\lambda$ are identical to
those described in §\ref{sec:exp2_protocol}.

\subsubsection{Evaluation criteria}

Validation is carried out using three quantitative indicators:
\begin{enumerate}
    \item \textit{Tracking ability:} Spearman rank correlation between
    $\lambda$ and the metric value, aggregated over methods and scales.
    \item \textit{Geometric specificity:} the coefficient $R_M(\lambda)$
    from formula~\eqref{eq:R}, computed as an increment relative to the
    ``clean'' output ($\lambda=0$).
    \item \textit{Architectural stability:} variability of
    $\Delta\operatorname{MSIQ}$ at the same $\lambda$ for the Classic
    and DNN groups.
\end{enumerate}

\subsubsection{Computational robustness test}

Special attention is paid to the influence of the moment discretization
method on the final score. Three implementations are compared:
\begin{enumerate}
    \item \texttt{skimage\_raw} --- the standard implementation based
    on the discrete grid;
    \item \texttt{pixel\_center\_delta} --- a model of masses concentrated
    at pixel centers;
    \item \texttt{pixel\_integrated\_constant} --- a model of
    piecewise-constant intensity with analytical integration of monomials
    over the pixel area.
\end{enumerate}
The results of this comparison are given in §\ref{sec:results_exp3}.

\subsection{Fourth experiment: validation on standard SISR benchmarks}

The aim of the fourth experiment is to test whether the geometric specificity
of MSIQ is preserved under a broader and more heterogeneous protocol than the
six illustrative images in Exp.~2--3.

The Set5, Set14, and Urban100 datasets are used. For each HR image $I_{GT}$,
LR input data are formed by bicubic downsampling with factors
$s\in\{2,3,4\}$. SR outputs are generated by classical methods
(\texttt{nearest}, \texttt{bilinear}, \texttt{bicubic}, \texttt{lanczos4})
and DNN models (\texttt{EDSR}, \texttt{ESPCN}, \texttt{FSRCNN},
\texttt{LapSRN}); EDSR, ESPCN, and FSRCNN are applied for scales
$\times 2,3,4$, while LapSRN is applied for scales $\times 2,4$.

The controlled degradation protocol (§\ref{sec:exp2_protocol}) is applied
to each SR output: four geometric degradations and the JPEG control at
$\lambda\in\{0,\,0.05,\,0.10,\,0.15,\,0.20\}$.
For each pair $(I_{GT},\,T_{\lambda}(I_{SR}))$, the following metrics are
computed:
$$
    \mathrm{PSNR},\quad
    \mathrm{SSIM},\quad
    \operatorname{MSIQ}_{RMSE},\quad
    \operatorname{MSIQ}_{W},\quad
    \mathrm{DISTS},\quad
    \mathrm{LPIPS},\quad
    \mathrm{ERQA}.
$$
The main criterion is the coefficient $R_M(\lambda)$
from formula~\eqref{eq:R}.

\subsection{General implementation parameters}

The software framework is implemented in Python~3.11 using
NumPy~1.26, OpenCV~4.9, and scikit-image~0.22. Moment computation is
based on \texttt{skimage.measure.moments}, followed by analytical
normalization according to formulas~\eqref{eq:nu}--\eqref{eq:descriptor}.
Before processing, all images are normalized to the range $[0,1]$.
The MSIQ parameters are fixed as follows: $N=4$,
$\Omega_N=\Omega_4^{\triangle}$, $|\Omega_4^{\mathrm{inf}}|=12$;
the main version is $\operatorname{MSIQ}_{RMSE}$, and the auxiliary
version is $\operatorname{MSIQ}_{W}$ with weights $w_{pq}=1/(1+p+q)$.

\section{Results}
\label{sec:results}

This section presents the results of the four experimental series and the
ablation analysis. The MSIQ results are compared with classical metrics
(PSNR, SSIM), perceptual metrics (LPIPS, DISTS), and the structure-oriented
ERQA~\cite{Lyapustin2022}.

\subsection{First experiment: validation of the scale-invariant property}
\label{sec:exp1}

\subsubsection{Computational stability of MSIQ in the discrete domain}

Both control conditions --- $\nu_{00}\approx 1$, $\nu_{10}\approx 0$,
$\nu_{01}\approx 0$, and $\operatorname{MSIQ}_{RMSE}(I,I)=0$ ---
were satisfied for all test images, confirming the absence of systematic
implementation error.

The statistics of $\operatorname{MSIQ}_{RMSE}$ values for scaled copies are
reported in Table~\ref{tab:exp1_msiq}. For smooth interpolators, MSIQ values
remain at the level of $10^{-7}$--$10^{-6}$, which corresponds to residual
discretization error. For the \texttt{nearest} method, the values are higher,
which is explained by the staircase artifacts of this interpolator; however,
they remain small in absolute scale.

\begin{table}[htbp]
\centering
\caption{Statistics of \(\operatorname{MSIQ}_{RMSE}\) values under
uniform scaling without forced resizing.}
\label{tab:exp1_msiq}
\begin{small}
\begin{tabular}{lcccc}
\hline
Interpolation method & Mean & Median & Min & Max \\
\hline
\texttt{area}
& \(6.66 \cdot 10^{-7}\)
& \(4.90 \cdot 10^{-7}\)
& \(1.40 \cdot 10^{-7}\)
& \(2.01 \cdot 10^{-6}\) \\
\texttt{bilinear}
& \(2.76 \cdot 10^{-6}\)
& \(6.72 \cdot 10^{-7}\)
& \(1.33 \cdot 10^{-7}\)
& \(4.42 \cdot 10^{-5}\) \\
\texttt{lanczos4}
& \(2.89 \cdot 10^{-6}\)
& \(3.39 \cdot 10^{-6}\)
& \(4.03 \cdot 10^{-7}\)
& \(7.05 \cdot 10^{-6}\) \\
\texttt{bicubic}
& \(2.97 \cdot 10^{-6}\)
& \(2.81 \cdot 10^{-6}\)
& \(3.69 \cdot 10^{-7}\)
& \(8.03 \cdot 10^{-6}\) \\
\texttt{nearest}
& \(3.13 \cdot 10^{-5}\)
& \(8.35 \cdot 10^{-6}\)
& \(1.97 \cdot 10^{-7}\)
& \(1.28 \cdot 10^{-4}\) \\
\hline
\end{tabular}
\end{small}
\end{table}

The increased mean value for \texttt{bilinear} ($2.76\cdot10^{-6}$)
is caused by a single outlier (max $4.42\cdot10^{-5}$), while the median
remains at the level of $6.72\cdot10^{-7}$. This outlier does not change
the main conclusion: under uniform scaling, MSIQ remains practically zero.

\subsubsection{Instability of PSNR and SSIM under forced resizing}

The aggregated results are reported in Table~\ref{tab:exp1_forced_resizing}.
The values of PSNR and SSIM substantially depend on the choice of the forced
resizing method. The mean finite PSNR spread is $45.48$~dB (median
$5.28$~dB), with a maximum of $653.83$~dB. The SSIM spread has mean
$0.044$ and maximum $0.249$. A total of 95 cases of infinite PSNR
(MSE$=0$) were recorded.

\begin{table}[htbp]
\centering
\caption{Instability of PSNR and SSIM caused by the \emph{forced resizing}
procedure. Values are grouped by the method of primary image scaling.}
\label{tab:exp1_forced_resizing}
\begin{small}
\begin{tabular}{lrrrrrr}
\hline
Scaling
& Mean
& Median
& Max
& Mean
& Max
& Inf \\
method
& $\Delta$PSNR
& $\Delta$PSNR
& $\Delta$PSNR
& $\Delta$SSIM
& $\Delta$SSIM
& PSNR \\
\hline
\texttt{lanczos4} & 78.06 & 19.80 & 645.90 & 0.0459 & 0.1253 & 9 \\
\texttt{bicubic}  & 75.42 & 11.16 & 653.83 & 0.0435 & 0.1192 & 15 \\
\texttt{bilinear} & 61.35 &  6.38 & 287.24 & 0.0420 & 0.1052 &  1 \\
\texttt{nearest}  &  3.68 &  2.48 &  11.70 & 0.0599 & 0.2495 & 35 \\
\texttt{area}     &  3.29 &  2.48 &   8.56 & 0.0284 & 0.0968 & 35 \\
\hline
\end{tabular}
\end{small}
\end{table}

Extreme values arise in degenerate cases where MSE approaches zero; therefore,
the main interpretation should rely on the median spread and the number of
Inf PSNR cases.

Thus, a substantial part of the numerical PSNR/SSIM result may reflect artifacts
of the external forced resizing procedure rather than the actual quality of
SR reconstruction. MSIQ is free from this dependence because it compares images
in the space of moment descriptors.

\subsection{Second experiment: geometric specificity of MSIQ}
\label{sec:results_exp2}

\subsubsection{Tracking ability}

Tracking ability was evaluated using Spearman rank correlation between
$\lambda$ and the metric response. For higher-is-better metrics
(PSNR, SSIM, ERQA), the sign of the correlation was inverted; a positive value
in Table~\ref{tab:exp2_tracking} indicates a correct monotonic response.

\begin{table}[htbp]
\centering
\caption{Second experiment: tracking ability for geometric degradations.
Mean expected signed Spearman correlation between $\lambda$ and the metric
response.}
\label{tab:exp2_tracking}
\begin{small}
\begin{tabular}{lc}
\hline
Metric & Mean expected $\rho$ \\
\hline
LPIPS                         & $0.904$ \\
DISTS                         & $0.841$ \\
$\operatorname{MSIQ}_{W}$     & $0.742$ \\
$\operatorname{MSIQ}_{RMSE}$  & $0.742$ \\
ERQA                          & $0.727$ \\
SSIM                          & $0.715$ \\
PSNR                          & $0.328$ \\
\hline
\end{tabular}
\end{small}
\end{table}

LPIPS and DISTS show the highest monotonicity, since they are constructed
in feature spaces that are sensitive to a broad range of visual changes.
Both versions of MSIQ have a stable positive correlation ($\rho\approx0.742$),
which confirms their ability to monotonically track increasing geometric
degradation. The low value for PSNR ($0.328$) is explained by the pixel-oriented
nature of the metric and its sensitivity to spatial alignment, rather than to
the geometric parameter $\lambda$.

\subsubsection{Geometric specificity: the geometric/JPEG ratio}

The key indicator is the coefficient $R_M(0.20)$ from formula~\eqref{eq:R}.
PSNR is excluded from Table~\ref{tab:exp2_specificity} because of its
instability with respect to pixel alignment, which prevents reliable computation
of $R_M$.

\begin{table}[htbp]
\centering
\caption{Second experiment: geometric specificity for $\lambda=0.20$.}
\label{tab:exp2_specificity}
\begin{small}
\begin{tabular}{lcc}
\hline
Metric & Geometric Mean / JPEG & Geometric Min / JPEG \\
\hline
$\mathbf{MSIQ_{RMSE}}$ & $\mathbf{71.51}$ & $9.91$ \\
$\mathbf{MSIQ_{W}}$    & $\mathbf{70.92}$ & $9.78$ \\
SSIM  & $4.85$ & $1.84$ \\
LPIPS & $3.86$ & $1.41$ \\
ERQA  & $3.22$ & $1.25$ \\
DISTS & $0.96$ & $0.44$ \\
\hline
\end{tabular}
\end{small}
\end{table}

MSIQ exceeds the JPEG response by approximately a factor of $71$ for both
versions. DISTS ($R_M\approx1$) practically does not distinguish geometric
degradation from JPEG degradation, despite having the highest tracking ability.
This result confirms that high monotonic tracking is a general perceptual
response, whereas $R_M$ measures geometric specificity. LPIPS and ERQA have
moderate specificity; SSIM is higher than learned perceptual metrics, but is
substantially inferior to MSIQ.

\subsubsection{Small-deformation regime ($\lambda=0.05$)}

\begin{table}[htbp]
\centering
\caption{Second experiment: geometric/JPEG specificity in the small-deformation
regime $\lambda=0.05$.}
\label{tab:exp2_small}
\begin{small}
\begin{tabular}{lc}
\hline
Metric & $R_M(0.05)$ \\
\hline
$\mathbf{MSIQ_{RMSE}}$ & $\mathbf{161.73}$ \\
$\mathbf{MSIQ_{W}}$    & $\mathbf{161.26}$ \\
SSIM  & $20.69$ \\
LPIPS & $11.47$ \\
ERQA  &  $5.36$ \\
DISTS &  $1.69$ \\
\hline
\end{tabular}
\end{small}
\end{table}

As the degradation strength decreases, the geometric specificity of MSIQ
increases to $R_M\approx161$, since the JPEG response remains negligible
whereas MSIQ already detects weak geometric violations. This property is
critically important for detecting subtle geometric anomalies in SR
reconstruction that are visually imperceptible.

\subsection{Third experiment: validation on the outputs of SISR algorithms}
\label{sec:results_exp3}

\subsubsection{Metric response under maximum degradation}

The mean metric values for $\lambda=0.20$ are reported in
Table~\ref{tab:exp3_max}. The ratio
$\operatorname{MSIQ}_{RMSE}(\text{geometric})$~/$\operatorname{MSIQ}_{RMSE}(\text{JPEG})$
ranges from $11.3$ (shear) to $29.0$ (perspective), indicating pronounced
geometric specificity even in the presence of SR artifacts.
ERQA decreases substantially for all geometric degradations, capturing the
destruction of edge structures. DISTS remains almost unchanged across all
types of degradation.

\begin{table}[htbp]
\centering
\caption{Third experiment: mean metric values on SISR outputs under
maximum degradation $\lambda=0.20$.}
\label{tab:exp3_max}
\small
\begin{tabular}{lccccc}
\hline
Degradation type
& $SSIM\uparrow$
& $\operatorname{MSIQ}_{RMSE}\downarrow$
& $DISTS\downarrow$
& $LPIPS\downarrow$
& $ERQA\uparrow$ \\
\hline
JPEG control & 0.792 & 0.000123 & 0.230 & 0.262 & 0.542 \\
Shear         & 0.512 & 0.001387 & 0.237 & 0.407 & 0.268 \\
Anis. affine  & 0.434 & 0.002465 & 0.261 & 0.494 & 0.211 \\
Rotation      & 0.391 & 0.002991 & 0.281 & 0.560 & 0.169 \\
Perspective   & 0.369 & 0.003568 & 0.259 & 0.535 & 0.192 \\
\hline
\end{tabular}
\end{table}

MSIQ and ERQA are complementary: the former measures changes in the global
moment-geometric description, while the latter measures the local correctness
of contours.

\subsubsection{Tracking ability on SISR outputs}

\begin{table}[htbp]
\centering
\caption{Third experiment: tracking ability on SISR outputs, measured by the
mean expected signed Spearman correlation.}
\label{tab:exp3_tracking}
\small
\begin{tabular}{lc}
\hline
Metric & Mean signed $\rho$ \\
\hline
$\mathbf{MSIQ_{W}}$    & $\mathbf{0.715}$ \\
$\mathbf{MSIQ_{RMSE}}$ & $\mathbf{0.713}$ \\
SSIM  & $0.649$ \\
LPIPS & $0.619$ \\
PSNR  & $0.615$ \\
ERQA  & $0.523$ \\
DISTS & $0.309$ \\
\hline
\end{tabular}
\end{table}

On SR outputs, both versions of MSIQ move to the top positions in terms of
tracking ability, whereas in Exp.~2 they were outperformed by LPIPS and DISTS.
This indicates that, after reconstruction, local texture features become a less
stable basis for tracking the geometric parameter, while the global moment
description preserves its sensitivity.

\subsubsection{Geometric specificity on SISR outputs}

\begin{table}[htbp]
\centering
\caption{Third experiment: geometric/JPEG ratio on SISR outputs at
$\lambda=0.20$.}
\label{tab:exp3_specificity}
\small
\begin{tabular}{lcc}
\hline
Metric & Geometric Mean / JPEG & Geometric Min / JPEG \\
\hline
$\mathbf{MSIQ_{RMSE}}$ & $\mathbf{122.07}$ & $-4.31^{*}$ \\
$\mathbf{MSIQ_{W}}$    & $\mathbf{121.08}$ & $-3.64^{*}$ \\
LPIPS & $47.17$ & $-1.95^{*}$ \\
ERQA  & $19.41$ & $2.25$ \\
SSIM  &  $8.75$ & $1.78$ \\
DISTS &  $1.66$ & $0.29$ \\
\hline
\multicolumn{3}{l}{\footnotesize $^{*}$ Negative minima occur in isolated
cases of local compensation in feature space.}
\end{tabular}
\end{table}

On SR outputs, MSIQ responds to geometric degradations more than 100 times
more strongly than to JPEG. DISTS ($R_M=1.66$) remains the least specific.
The negative minima for MSIQ and LPIPS are isolated artifacts and do not affect
the general conclusion.

\subsubsection{Architectural stability}

\begin{table}[htbp]
\centering
\caption{Third experiment: architectural stability of MSIQ at
$\lambda=0.20$.
$\Delta\operatorname{MSIQ}=\operatorname{MSIQ}(0.20)-\operatorname{MSIQ}(0)$.}
\label{tab:exp3_arch_stability}
\small
\begin{tabular}{lrrrrrr}
\hline
Degradation type
& $\Delta$RMSE Classical
& $\Delta$RMSE DNN
& Diff.,\%
& $\Delta W$ Classical
& $\Delta W$ DNN
& Diff.,\% \\
\hline
Shear            & 0.001347 & 0.001250 & 7.14 & 0.003324 & 0.003086 & 7.14 \\
Anisotropic aff. & 0.002442 & 0.002320 & 5.01 & 0.006003 & 0.005692 & 5.19 \\
Rotation         & 0.002975 & 0.002841 & 4.53 & 0.007345 & 0.007006 & 4.61 \\
Perspective      & 0.003564 & 0.003412 & 4.28 & 0.008626 & 0.008262 & 4.22 \\
\hline
Mean             & 0.002582 & 0.002456 & 4.90 & 0.006324 & 0.006012 & 4.95 \\
\hline
\end{tabular}
\end{table}

The mean relative difference between the Classical and DNN groups is
$4.90\%$ for $\operatorname{MSIQ}_{RMSE}$ and $4.95\%$ for
$\operatorname{MSIQ}_{W}$. The largest difference, shear ($7.14\%$),
remains small compared with the absolute excess over the JPEG control.
The analysis of the increment $\Delta\operatorname{MSIQ}$ confirms that the
metric primarily measures the introduced geometric degradation, rather than
the architectural class of the SR algorithm.

\subsubsection{Small-deformation regime ($\lambda=0.05$)}

\begin{table}[htbp]
\centering
\caption{Third experiment: geometric/JPEG specificity at $\lambda=0.05$.}
\label{tab:exp3_small}
\small
\begin{tabular}{lc}
\hline
Metric & $R_M(0.05)$ \\
\hline
$\mathbf{MSIQ_{RMSE}}$ & $\mathbf{386.67}$ \\
$\mathbf{MSIQ_{W}}$    & $\mathbf{378.56}$ \\
ERQA  & $62.40$ \\
SSIM  & $52.52$ \\
LPIPS & $-11.17$ \\
DISTS & $-24.58$ \\
\hline
\end{tabular}
\end{table}

The negative values for LPIPS and DISTS arise because
$\Delta M_{\mathrm{JPEG}}<0$: weak JPEG compression is interpreted by these
metrics as an ``improvement'' relative to the undegraded SR output.
MSIQ is free from this effect: its response to JPEG remains small and
consistently positive under all conditions.

The comparative test of the three implementations of moment discretization
(\texttt{skimage\_raw}, \texttt{pixel\_center\_delta},
\texttt{pixel\_integrated\_constant}) 
produced practically identical values of \(\operatorname{MSIQ}_{RMSE}\) and
\(R_M(\lambda)\); therefore, \texttt{skimage\_raw} is used hereafter as the
standard baseline implementation.

\subsection{Fourth experiment: validation on standard SISR benchmarks}
\label{subsec:results_experiment4_standard_datasets}

The fourth experiment covered all images from Set5, Set14, and Urban100
for scales $\times2,\times3,\times4$.

The mean metric values at $\lambda=0.20$ are reported in
Table~\ref{tab:e4_severity_lambda_max}. JPEG compression remains the least
destructive influence: $\operatorname{MSIQ}_{RMSE}=0.000315$, whereas for
geometric degradations the values increase by an order of magnitude or more.
The strongest MSIQ response is produced by rotation and anisotropic affine
degradation, which is consistent with their effect on the global moment
structure.

\begin{table}[ht]
\centering
\caption{Experiment 4: mean metric values at $\lambda=0.20$.}
\label{tab:e4_severity_lambda_max}
\resizebox{\textwidth}{!}{%
\begin{tabular}{lrrrrrrr}
\hline
Degradation type
& PSNR & SSIM
& $\operatorname{MSIQ}_{RMSE}$
& $\operatorname{MSIQ}_{W}$
& DISTS & LPIPS & ERQA \\
\hline
JPEG               & 26.0991 & 0.7835 & 0.000315 & 0.000325 & 0.2202 & 0.2402 & 0.6606 \\
Perspective        & 13.8762 & 0.3291 & 0.003658 & 0.003775 & 0.2289 & 0.4348 & 0.3478 \\
Shear              & 13.4374 & 0.3475 & 0.004952 & 0.005251 & 0.2588 & 0.4704 & 0.3219 \\
Anisotropic affine & 12.6114 & 0.2907 & 0.006034 & 0.006305 & 0.2671 & 0.5515 & 0.2906 \\
Rotation           & 12.1776 & 0.2925 & 0.006974 & 0.007429 & 0.2734 & 0.5529 & 0.2841 \\
\hline
\end{tabular}
}
\end{table}

The geometric/JPEG separation indicators are reported in
Table~\ref{tab:e4_geometric_jpeg}. At $\lambda=0.20$, the value of $R_M$
for MSIQ is $100.49$ for $\operatorname{MSIQ}_{RMSE}$ and $101.18$ for
$\operatorname{MSIQ}_{W}$. In the small-deformation regime
($\lambda=0.05$), the effect increases to $428.51$ and $443.49$,
respectively.

\begin{table}[ht]
\centering
\caption{Experiment 4: delta-based geometric/JPEG separation for
$\lambda=0.20$ and $\lambda=0.05$.}
\label{tab:e4_geometric_jpeg}
\resizebox{\textwidth}{!}{%
\begin{tabular}{lrrrr}
\hline
Metric
& $R_M(0.20)$
& Mean geometric $\Delta M$, $\lambda=0.20$
& Mean JPEG $\Delta M$, $\lambda=0.20$
& $R_M(0.05)$ \\
\hline
$\operatorname{MSIQ}_{RMSE}$ & 100.49 & 0.005143 & 0.000051 & 428.51 \\
$\operatorname{MSIQ}_{W}$    & 101.18 & 0.005420 & 0.000054 & 443.49 \\
DISTS & 2.05  & 0.064895 & 0.031610 & $-15.52$ \\
LPIPS & 34.05 & 0.263496 & 0.007739 & $-27.02$ \\
SSIM  & 12.40 & 0.511808 & 0.041280 & 67.24 \\
ERQA  & 21.95 & 0.365884 & 0.016672 & 111.63 \\
PSNR  & 15.55 & 14.013000 & 0.901024 & 68.22 \\
\hline
\end{tabular}
}
\end{table}

Compared with Exp.~3, $R_M(0.20)$ for LPIPS decreases from $47.17$ to
$34.05$, which reflects the greater diversity of SISR outputs and the
sensitivity of LPIPS to local texture properties of the sample. MSIQ preserves
$R_M\approx100$ independently of the dataset.

For DISTS and LPIPS, the value of $R_M(0.05)$ becomes negative: the mean JPEG
response takes a negative sign, so weak compression is interpreted as an
improvement. MSIQ and ERQA do not exhibit this effect.

Tracking ability for all metrics is reported in Table~\ref{tab:e4_tracking}.
The median Spearman correlation is $1.000$ for each metric, indicating that
in most trajectories any metric perfectly ranks the degradation levels.
Tracking ability by itself does not differentiate the metrics; the decisive
criterion is geometric/JPEG separation.

\begin{table}[ht]
\centering
\caption{Experiment 4: tracking ability for geometric degradations.}
\label{tab:e4_tracking}
\begin{tabular}{lrr}
\hline
Metric & Mean $\rho$ & Median $\rho$ \\
\hline
LPIPS                        & 0.9788 & 1.0000 \\
PSNR                         & 0.9780 & 1.0000 \\
$\operatorname{MSIQ}_{RMSE}$ & 0.9778 & 1.0000 \\
$\operatorname{MSIQ}_{W}$    & 0.9765 & 1.0000 \\
SSIM                         & 0.9723 & 1.0000 \\
ERQA                         & 0.9324 & 1.0000 \\
DISTS                        & 0.9173 & 1.0000 \\
\hline
\end{tabular}
\end{table}

Thus, the fourth experiment confirms that the geometric specificity of MSIQ is
preserved on standard SISR benchmarks: the properties observed in Exp.~2--3
remain stable on a broader and more heterogeneous sample.

\subsection{Ablation analysis}
\label{subsec:ablation}

This subsection studies the influence of the maximum moment order $N$ on the
numerical stability of $\operatorname{MSIQ}_{RMSE}$ and its agreement with
basic FR metrics.

The ablation was conducted for $N\in\{3,4,\ldots,12\}$; the presence of
\texttt{NaN}/\texttt{Inf}, the mean absolute value of the components
$\overline{|\nu_{pq}|}$, and rank correlations between
$\operatorname{MSIQ}_{RMSE}$ and PSNR/SSIM were monitored. The negative sign
of the correlations is expected, since MSIQ is a degradation measure.

\begin{table}[H]
\centering
\caption{Ablation by order $N$: numerical stability and correlation of
$\operatorname{MSIQ}_{RMSE}$ with PSNR and SSIM.}
\label{tab:ablation}
\begin{tabular}{c|ccc|cc}
\hline
$N$
& \texttt{NaN} & \texttt{Inf} & $\overline{|\nu_{pq}|}$
& $\rho(\mathrm{PSNR},\operatorname{MSIQ}_{RMSE})$
& $\rho(\mathrm{SSIM},\operatorname{MSIQ}_{RMSE})$ \\
\hline
3  & No & No & 0.091380 & $-0.581657$ & $-0.559682$ \\
4  & No & No & 0.065242 & $-0.566801$ & $-0.542832$ \\
5  & No & No & 0.046110 & $-0.588028$ & $-0.557174$ \\
6  & No & No & 0.035586 & $-0.592958$ & $-0.561161$ \\
7  & No & No & 0.027567 & $-0.604292$ & $-0.577175$ \\
8  & No & No & 0.022366 & $-0.622401$ & $-0.587594$ \\
9  & No & No & 0.018265 & $-0.640074$ & $-0.600360$ \\
10 & No & No & 0.015340 & $-0.649296$ & $-0.604733$ \\
11 & No & No & 0.012967 & $-0.663380$ & $-0.612998$ \\
12 & No & No & 0.011167 & $-0.680114$ & $-0.630233$ \\
\hline
\end{tabular}
\end{table}

No numerical anomalies arise in the range $N=3,\ldots,12$.
As $N$ increases, $\overline{|\nu_{pq}|}$ decreases as a consequence of
normalization, while the correlation with PSNR/SSIM increases in absolute
value because higher-order moments add sensitivity to details. The selected
value $N=4$ is a practical compromise between informativeness, numerical
stability, and ease of interpretation.

Both versions of MSIQ exhibit qualitatively identical behavior in all
experiments; $\operatorname{MSIQ}_{W}$ acts as a robust modification with a
reduced contribution of higher-order moments, whereas
$\operatorname{MSIQ}_{RMSE}$ is the baseline unweighted version. The main
conclusions of the paper do not depend on the choice between them.

\section{Discussion}
\label{sec:discussion}

\subsection{The place of MSIQ among SISR quality assessment metrics}

MSIQ is not another universal image quality metric. It measures one specific
aspect: to what extent $I_{SR}$ preserves the scale-invariant
moment-geometric structure of the reference image $I_{GT}$. The experimental
results make it possible to clearly distinguish the niche of MSIQ among related
tools.

PSNR and SSIM measure signal or local structural similarity in the pixel
representation. In the case of images with different sizes, their computation
requires forced resizing, which introduces an error into the comparison that is
independent of the quality of the SR model; this effect is quantitatively
documented in Exp.~1 (Table~\ref{tab:exp1_forced_resizing}). LPIPS and DISTS
approximate perceptual similarity through feature spaces of pretrained networks,
while ERQA analyzes local edge structure without neural-network training.
MSIQ complements this set by assessing global moment-geometric consistency:
mass distribution, centroid, and higher-order moments. All these metrics answer
different questions and are therefore complementary rather than competing.

An important consequence of the first experiment is that the scale-free nature
of MSIQ is not reduced merely to the technical possibility of comparing images
of different sizes. Even when $I_{GT}$ and $I_{SR}$ have the same pixel size,
which is the standard situation in SISR benchmarks, MSIQ evaluates a different
type of correspondence: not pixel-wise error, but preservation of a
scale-invariant moment description. Thus, the advantage of the resizing-free
approach lies not only in eliminating dependence on forced resizing, but also
in a fundamentally different level of image representation.

\subsection{Tracking ability and geometric specificity as different properties}

One of the key methodological results of this work is the empirical confirmation
that the ability of a metric to monotonically track increasing degradation
strength, or tracking ability, and the ability to distinguish geometric
degradation from a non-geometric artifact, or geometric specificity, are
independent properties. A metric with the highest tracking ability may have
minimal geometric specificity.

This is exactly how LPIPS and DISTS behave. In Exp.~2, they show the highest
rank correlations with the parameter $\lambda$ ($\rho=0.904$ and $0.841$,
respectively), but their geometric/JPEG separation coefficient $R_M(0.20)$ is
$3.86$ and $0.96$, respectively; in fact, they do not distinguish geometric
distortion from JPEG compression. The reason is that the feature spaces of deep
networks, being sensitive to a broad spectrum of local changes, respond to any
degradation rather than selectively to geometry. The instability of these
metrics in the small-deformation regime ($\lambda=0.05$: negative $R_M$ values
due to an ``improvement'' under JPEG) confirms this non-directional sensitivity.

MSIQ has lower tracking ability ($\rho\approx0.742$ in Exp.~2), but much higher
geometric specificity ($R_M\approx71$--$161$), which increases as the
degradation strength decreases. This agrees with the mathematical nature of the
measure: normalized moments are insensitive to uniform scaling and translation,
but sensitive to anisotropic deformations, shear, and perspective distortions.

\subsection{MSIQ, ERQA, and learned perceptual metrics}

The comparison between MSIQ and ERQA is particularly informative because both
metrics do not use pretrained neural networks. However, they describe different
levels of geometric structure. ERQA is oriented toward the local structure of
edges: it responds to shifts, deformations, or loss of contours. MSIQ describes
global structure through normalized moments: mass distribution and its spatial
organization as a whole.

In the small-deformation regime ($\lambda=0.05$), ERQA shows $R_M=62.4$
(Exp.~3) and $111.6$ (Exp.~4), substantially higher than LPIPS and DISTS, but
lower than MSIQ ($386.7$ and $428.5$, respectively). This is explained by the
fact that weak shear or rotation immediately shifts the positions of contours,
whereas moderate JPEG compression preserves the main edge structure. Thus,
ERQA is an effective edge-based detector of local geometric violations, whereas
MSIQ is a detector of global moment-geometric deviations. These two metrics are
complementary: an image may preserve local edges while having distorted
proportions, or conversely, it may contain local edge artifacts while preserving
global geometry.

Learned perceptual metrics occupy yet another position. The results are
consistent with the theoretically justified perception-distortion
tradeoff~\cite{Blau2018}: maximizing perceptual similarity does not guarantee
the preservation of geometric structure. The tendency of such metrics toward
texture bias~\cite{Zhang2018} means that they may ignore violations of global
geometry if the texture and contrast profiles are preserved. Thus, perceptual
closeness and geometric correctness are different, complementary aspects of
SISR quality.

\subsection{Practical recommendations}

The results of this work make it possible to formulate a concrete protocol for
tasks in which geometric correctness is an independent requirement. It is
advisable to use a combined set of metrics: PSNR/SSIM for baseline fidelity
assessment; LPIPS or DISTS for perceptual naturalness; ERQA for local edge
structure; and MSIQ for global moment-geometric consistency. Such a set covers
four fundamentally different aspects of SR reconstruction quality.

MSIQ is especially relevant for tasks in which geometric error has practical
consequences: technical inspection, medical imaging, remote sensing,
cartographic images, and architectural images. In these domains, an SR model
must preserve proportions and the spatial organization of objects, rather than
merely reproduce convincing texture. For tasks where only perceptual appeal or
texture detail is important, MSIQ is not the primary criterion.

\subsection{Limitations}
\label{sec:limitations}

\textit{Global nature of the assessment.}
As an integral measure, MSIQ responds weakly to spatially localized artifacts
that do not shift the overall moment descriptor. In tasks where the target
object occupies only a small part of the frame, the use of an ROI mask or a
local, window-based implementation of MSIQ is recommended.

\textit{Photometric non-invariance.}
Normalized central moments are not invariant to global photometric
transformations such as brightness, contrast, or gamma changes. If such
differences are not considered degradation, preliminary histogram normalization
or contrast matching is required before computing MSIQ.

\textit{Absence of a semantic dimension.}
MSIQ assesses the structural integrity of the mass distribution, but it is not
a semantic metric. Hallucinations of generative models, that is, the appearance
of plausible but fictitious details, may not affect the global moment descriptor
if they do not change the overall spatial organization of the image. To detect
semantic artifacts, MSIQ must be complemented by perceptual or content-aware
metrics.

\textit{Limited validation on modern generative models.}
The current validation covers classical interpolators and CNN-based SR models
(FSRCNN, ESPCN, LapSRN, EDSR). At the same time, one of the main motivations
of the Introduction is the spread of generative and diffusion-based SR
approaches (SwinIR, Real-ESRGAN, StableSR, SeeSR), which are prone to specific
structural deformations. The behavior of MSIQ when evaluating such models,
in particular whether geometric specificity and tracking ability remain stable
on their outputs, remains an open question and is a priority direction for
future research.

\textit{Absence of correlation with subjective scores.}
The relationship between MSIQ and subjective quality scores (MOS/DMOS) has not
been investigated. Since MSIQ is not a perceptual metric, high correlation with
MOS is not its goal. However, understanding the structure of this correlation,
for example, for which classes of degradations MSIQ and perceptual metrics
diverge in their assessments, is important for positioning MSIQ within existing
IQA taxonomies.

\section{Conclusions}
\label{sec:conclusion}

This paper proposes a new measure, MSIQ (Moment-based Scale-Invariant
Quality), for assessing the geometric correctness of single image
super-resolution results. Unlike standard FR metrics, MSIQ compares
$I_{GT}$ and $I_{SR}$ in the space of normalized central geometric
moments, which are scale-invariant descriptors that do not require the
images to be brought to a common pixel size.

The proposed measure is:
\begin{enumerate}
\item \emph{geometry-specific} --- it responds primarily to changes in
    the scale-invariant moment structure, rather than to general
    perceptual degradation;
    \item \emph{resizing-free} --- the comparison is performed without
    forced resizing, which removes the dependence of the evaluation on
    an external interpolation procedure;
    \item \emph{model-free} --- it is defined analytically through
    normalized central geometric moments and does not require a
    pretrained neural network.
\end{enumerate}

The first experiment confirmed the practical stability of MSIQ under
uniform scaling: the residual error was $10^{-7}$--$10^{-6}$ for smooth
interpolators, whereas the median PSNR spread caused by the choice of
the forced resizing method was $5.28$~dB; individual degenerate cases
with \(MSE\approx 0\) led to extreme PSNR values. In controlled
experiments, MSIQ showed the clearest separation of geometric
deformations from the JPEG control: the geometric/JPEG separation was
$R_M(0.20)\approx71$ on model distortions, $R_M\approx122$ on SISR
outputs, and $R_M\approx100$ on the standard Set5, Set14, and Urban100
benchmarks. In the small-deformation regime \((\lambda=0.05)\), this
effect becomes stronger, especially on SISR outputs and standard
benchmarks, where \(R_M\) exceeds \(380\), confirming the suitability of
MSIQ for diagnosing visually imperceptible geometric anomalies.

At the same time, MSIQ does not replace PSNR, SSIM, LPIPS, DISTS, or
ERQA, but complements them. Together, these metrics cover four different
aspects of quality: signal fidelity, perceptual naturalness, local edge
structure, and global moment-geometric consistency. The recommended
protocol is to use them jointly in tasks where geometric correctness is
an independent requirement.

The main limitations of the measure are related to its global nature
(sensitivity to background and weak response to local artifacts), the
absence of photometric invariance, and limited validation on modern
generative SR models (§\ref{sec:limitations}).

Future research includes several directions.
First, it is necessary to investigate the correlation of MSIQ with
subjective quality scores, MOS/DMOS, on available SISR-IQA datasets.
Although MSIQ is not a perceptual metric, such validation would make it
possible to calibrate its practical interpretation and to determine for
which classes of degradations it complements or diverges from human
judgments.

Second, the comparative empirical validation of MSIQ should be broadened
by including direct comparisons with closely related geometry-disparate
full-reference metrics, in particular DeepSSIM. Such a comparison would
make it possible to position MSIQ more precisely with respect to modern
approaches that also avoid direct pixel-wise comparison or are robust to
geometric mismatch between the reference and test images.

Third, it is advisable to move from Euclidean distances to
covariance-aware measures, in particular Mahalanobis distance or
whitened Euclidean distance in the space of moment descriptors, including
descriptors constructed from orthogonal moments, which may improve the
calibration of MSIQ for heterogeneous classes of images.

Fourth, the development of a local, window-based version of MSIQ would
make it possible to detect spatially localized geometric artifacts that
are invisible in the global moment description.

Fifth, and this is the highest-priority direction, MSIQ should be
validated on modern generative and diffusion-based SR models, such as
SwinIR, Real-ESRGAN, StableSR, SeeSR, and related approaches. It is
precisely for these models that the problem of structural deformations is
most relevant; therefore, confirming the geometric specificity and
tracking ability of MSIQ on their outputs is a key condition for the
practical applicability of the measure.

\section*{Code availability}
The source code of the MSIQ metric is open and available in the GitHub
repository\footnote{\text{\url{https://github.com/LeonidBed/msiq-metrics}}}.
MSIQ has also been released as a Python package
\[
    \texttt{ msiq-metrics}
\]
The package provides tools for computing normalized central moments and
both variants of the metric
($\operatorname{MSIQ}_{RMSE}$, $\operatorname{MSIQ}_{W}$) for
monochrome and color images.

\end{document}